\pdfoutput=1

\documentclass[11pt]{article}

\usepackage[preprint]{ACL2025}

\usepackage{times}
\usepackage{latexsym}

\usepackage[T1]{fontenc}

\usepackage[utf8]{inputenc}

\usepackage{microtype}

\usepackage{inconsolata}

\usepackage{graphicx}
\usepackage{enumitem}
\usepackage[ruled, linesnumbered, noend, vlined]{algorithm2e}

\usepackage{amssymb}
\usepackage{amsmath}
\usepackage{amsthm}
\usepackage{subfigure}
\usepackage{booktabs}
\usepackage{placeins}
\usepackage{multirow}
\usepackage{makecell}
\usepackage{framed}

\usepackage{numprint}
\npthousandsep{,}
\npdecimalsign{.}

\newtheorem{problem}{Problem}

\usepackage[most]{tcolorbox}
\usepackage{tablefootnote}

\newcommand{\hide}[1]{}

\usepackage{pifont}

\title{AppealCase: A Dataset and Benchmark for Civil Case Appeal Scenarios}

\author{Yuting Huang$^1$, Meitong Guo$^1$, Yiquan Wu$^{1*}$, Ang Li$^1$\\
\textbf{Xiaozhong Liu$^2$, Keting Yin$^1$, Changlong Sun$^3$, Fei Wu$^1$, Kun Kuang$^{1*}$} \vspace{0.5em}\\
\small $^1$Zhejiang University, Hangzhou, China \\
\small $^2$Worcester Polytechnic Institute, Worcester, USA \\
\small $^3$Alibaba Group, Hangzhou, China \\
\small\texttt{\{yutinghuang, wuyiquan, leeyon, yinkt, wufei, kunkuang\}@zju.edu.cn} \\
\small\texttt{guomeitong2003@163.com}, \texttt{xliu14@wpi.edu}, \texttt{changlong.scl@taobao.com} \\
}

\begin{document}
\maketitle

\renewcommand{\thefootnote}{\fnsymbol{footnote}}
\footnotetext[1]{Corresponding author.}

\begin{abstract}
Recent advances in LegalAI have primarily focused on individual case judgment analysis, often overlooking the critical appellate process within the judicial system. Appeals serve as a core mechanism for error correction and ensuring fair trials, making them highly significant both in practice and in research. To address this gap, we present the AppealCase dataset, consisting of 10,000 pairs of real-world, matched first-instance and second-instance documents across 91 categories of civil cases. The dataset also includes detailed annotations along five dimensions central to appellate review: judgment reversals, reversal reasons, cited legal provisions, claim-level decisions, and whether there is new information in the second instance. Based on these annotations, we propose five novel LegalAI tasks and conduct a comprehensive evaluation across 20 mainstream models. Experimental results reveal that all current models achieve less than 50\% F1 scores on the judgment reversal prediction task, highlighting the complexity and challenge of the appeal scenario. We hope that the AppealCase dataset will spur further research in LegalAI for appellate case analysis and contribute to improving consistency in judicial decision-making.
\end{abstract}

\section{Introduction}

The development of Legal Artificial Intelligence (LegalAI) has undergone an evolution from rule-based systems and formal logic reasoning \cite{sergot1986british}, to statistical learning models \cite{chalkidis-etal-2020-legal}, and further to large language models (LLMs) \cite{zhou2024lawgpt}. In particular, large language models have significantly expanded the boundaries of their applications in the legal field. General-purpose large language models have outperformed previous fine-tuned models in several core tasks, such as legal consultation Q\&A \cite{buttner-habernal-2024-answering} and similar case retrieval \cite{wiratunga2024cbr}.

However, current LegalAI research, in tasks such as legal provision recommendation \cite{zheng2022lawrec}, legal judgment prediction \cite{tong2024legal}, and court view generation \cite{li-etal-2024-enhancing}, mainly focuses on one-shot judgment cases and generally neglects the appeal procedure, which is a crucial and institutionally significant legal process. As a result, there remains a gap in modeling the appellate judgment process.

\begin{figure}[t]
    \centering
    \includegraphics[width=\columnwidth]{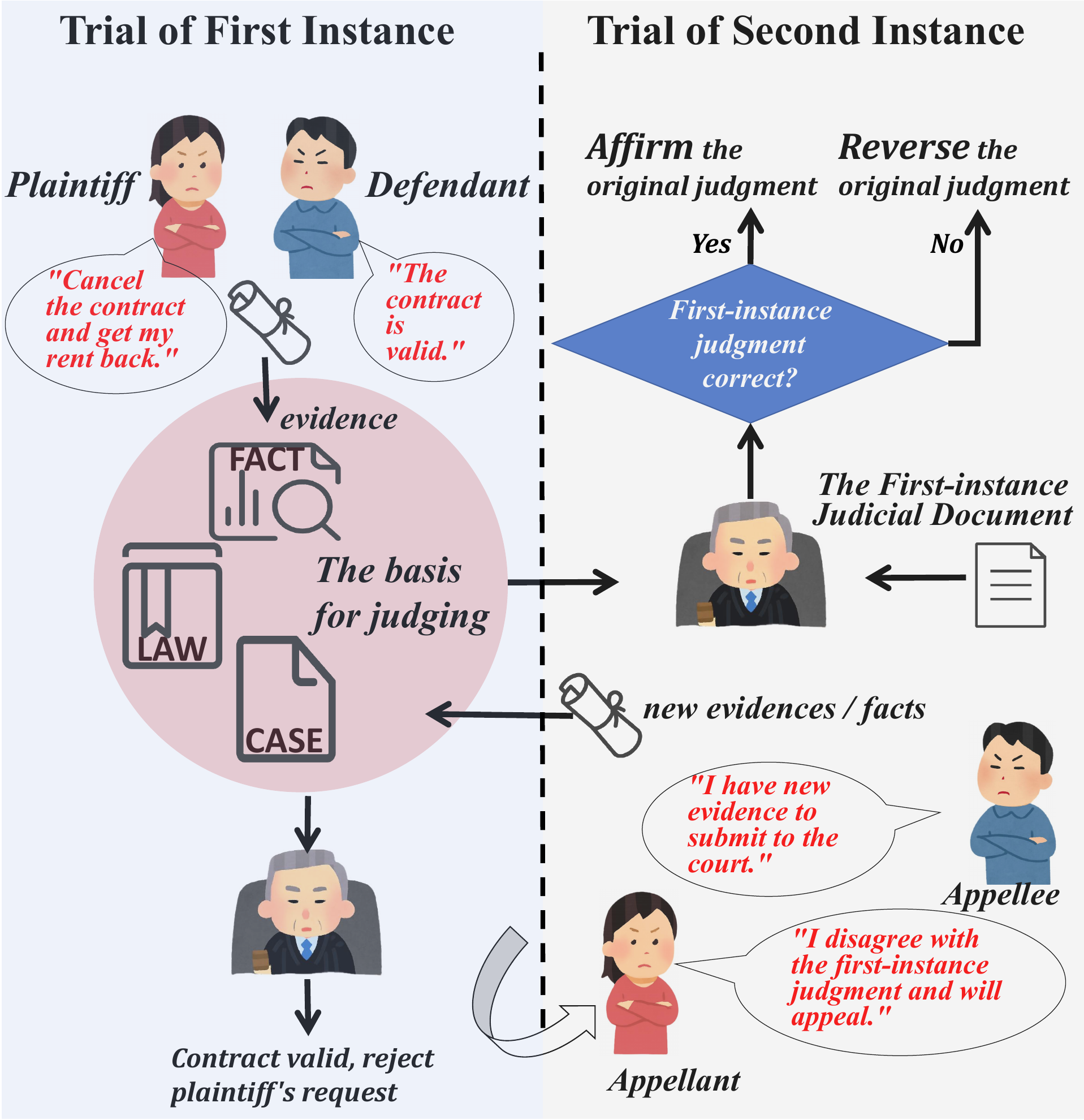}
    \caption{Procedural flow from first-instance trial to second-instance judgment, illustrating the roles of appellants, new evidence submission, and possible outcomes (affirmation or reversal) of the original  judgment.}
    \label{fig:fig1}
\end{figure}

Appellate courts are essential to ensuring consistency and correctness in the legal system in both civil-law and common-law traditions \cite{lax2007constructing}. As shown in Figure \ref{fig:fig1}, when parties to a case are dissatisfied with the outcome of the original judgment, they may file an appeal, seeking correction from the appellate court. While common-law appellate courts typically focus on reviewing legal errors and defer to the trial court’s findings of fact \cite{merryman2018civil}, appellate courts in civil-law jurisdictions often conduct a more comprehensive review, including reassessment of factual determinations \cite{seidman2016new}. Despite these procedural differences, both systems rely on appellate review as a critical mechanism for upholding legal integrity and ensuring uniformity across cases.

The first-instance court conducts the initial trial of a case, aiming to comprehensively ascertain the facts and apply the law to render a judgment. In contrast, the second-instance court is limited to the claims raised in the appeal and, taking into account new evidence and other information submitted during the appellate proceedings, focuses on reviewing whether there were factual or legal errors in the first-instance judgment. The appeal system is an important judicial error-correction mechanism, providing parties with opportunities to correct mistakes and obtain remedies. It differs from the first-instance scenario in both function and mechanism. Despite its practical and theoretical importance, research on LegalAI in the appellate scenario is  largely remains unexplored.

To fill this gap, in this paper, we construct the AppealCase dataset, collecting \numprint{10000} pairs of matched first-instance and second-instance judgment documents from China Judgments Online~\footnote{\url{https://wenshu.court.gov.cn/}}, covering 91 civil causes of action, with 50\% of the cases resulting in judgment reversals. We design a dedicated annotation scheme for the appeal scenario and complete the structuring and annotation of the judgment documents.

Based on the AppealCase dataset, we also propose several new tasks. Judgment Reversal Prediction can help first-instance courts avoid erroneous judgments. Legal Provision Recommendation and Judgment Prediction tasks can assist second-instance courts in reviewing and adjudicating appeal cases, while Court View Generation can help second-instance courts in drafting judgment documents.

We conduct a comprehensive evaluation of LLMs in the appeal scenario, validating 20 non-reasoning, reasoning, and domain-specific models on the newly proposed tasks. The results show that existing LLMs perform poorly on these new tasks in the appellate scenario, with all models achieving an average F1 score of less than 50\% on the judgment reversal prediction task. This highlights the research value of the AppealCase dataset and the newly proposed LegalAI tasks.

The contributions of this paper can be summarized as follows:

\begin{enumerate}[itemsep=3pt,topsep=3pt,parsep=0pt]
\item We investigate the LegalAI task from the perspective of appellate court.

\item We propose a dataset, AppealCase, which contains 10,000 pairs of matched first-instance and second-instance documents, covering 91 types of civil case causes.

\item We set up new LegalAI tasks and conducted thorough evaluations on 20 models, verifying that current LLMs are unable to handle the appeal scenario effectively.

\item The dataset will be made publicly available under the CC BY-NC 4.0 License at \url{https://github.com/ythuang02/AppealCase/}.

\end{enumerate}

\section{The AppealCase Dataset for Appellate Case Analysis}

\begin{figure*}[t]
    \centering
    \includegraphics[width=\textwidth]{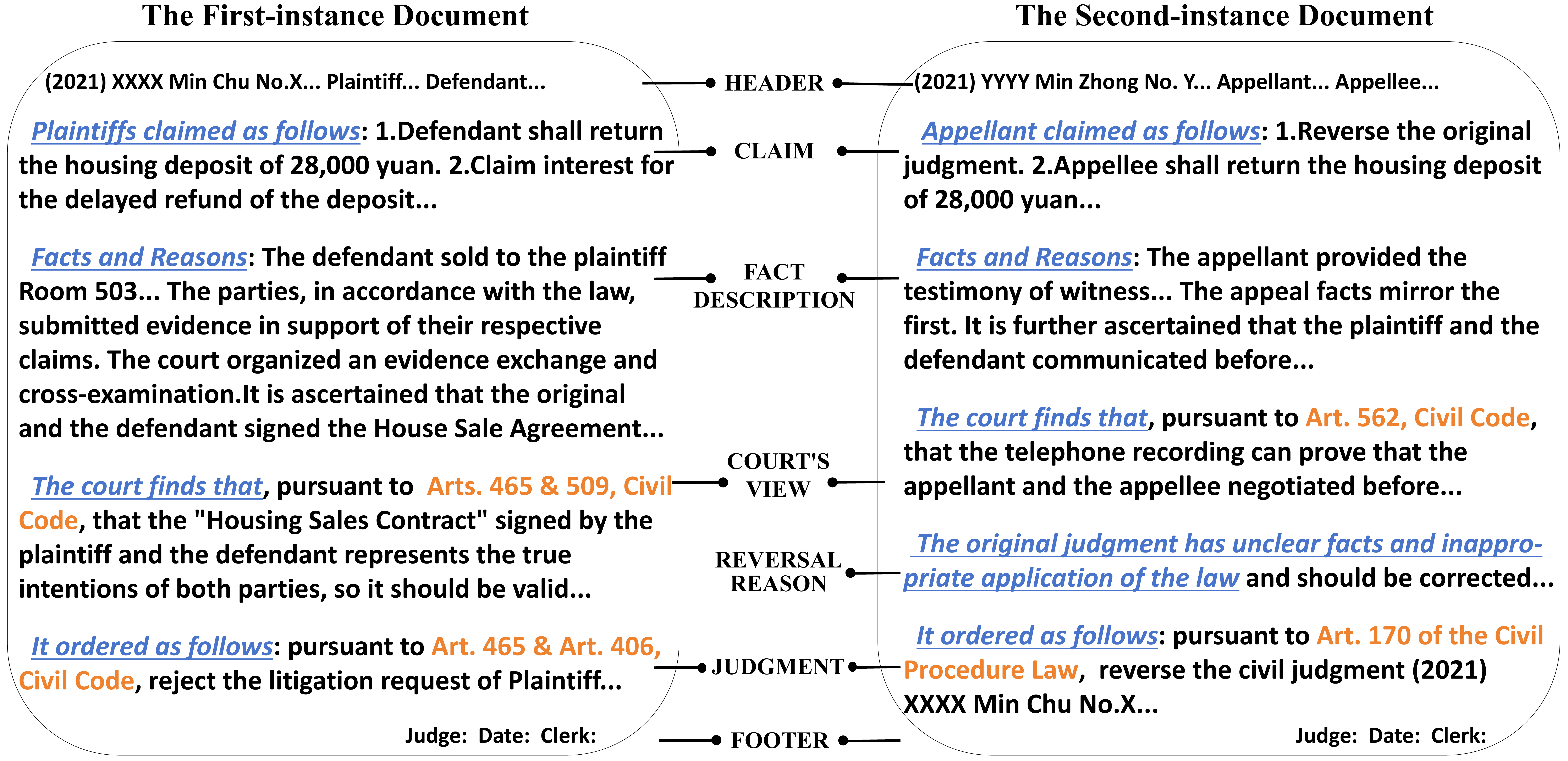}
    \caption{A comparative example of first-instance and second-instance documents. The diagram illustrates the structural correspondence between sections such as claims, facts, court’s view, and judgment. Underlined phrases represent typical legal expressions used to segment each paragraph in real-world documents.}
    \label{fig:document}
\end{figure*}

A standard judgment document in first-instance or second-instance cases, as illustrated in Figure \ref{fig:document}, typically includes the following sections: the \textbf{header}, which contains metadata such as the case number and court name; the \textbf{claim}, which summarizes the plaintiff’s or appellant's demands; the \textbf{fact description}, presenting the court’s account of the case facts, including statements from both parties and relevant evidence; the \textbf{court’s view}, which explains the legal reasoning and application of law. In appeal cases where the original judgment is modified, an additional section detailing the reasons for reversal is included to justify the appellate court’s disagreement with the lower court. The document concludes with the \textbf{judgment}, stating the final decision, and the \textbf{footer}, which provides the names of the judges and judgment date. We conducted further background analysis in Appendix \ref{appendix:background}.

\subsection{Dataset Overview and Statistics}

AppealCase contains \numprint{10000} cases, covering 91 types of civil causes. Each case includes a pair of matched first-instance and second-instance judgments, as shown in Figure 2, as well as five annotations.

\begin{table}[t]
\centering
\footnotesize
\renewcommand{\arraystretch}{1.00}
\scalebox{0.85}{
\begin{tabular}{lc}
\toprule
\textbf{Cause of Action} & \textbf{Proportion} \\
\toprule
Private Lending & 13.08\% \\
Labor Dispute & 9.94\% \\
Sales Contract & 9.86\% \\
Motor Vehicle Traffic Accident & 6.36\% \\
Contract & 5.28\% \\
Housing Lease Contract & 4.20\% \\
Construction Contract & 3.50\% \\
Labor Contract & 3.20\% \\
Housing Sale Contract & 2.74\% \\
Lease Contract & 2.18\% \\
\bottomrule
\end{tabular}
}
\caption{Proportion of the top-10 causes of action.}\label{tab:statistic_casecause}
\end{table}

\begin{table}[t]
\footnotesize
\centering
\renewcommand{\arraystretch}{1.0}
\scalebox{0.92}{
\begin{tabular}{l|c}
\toprule
\textbf{Type} & \textbf{Dataset} \\
\midrule
\# Cases & \numprint{10000} \\
\# Types of Cause of Actions & \numprint{91} \\
\midrule
Avg. Number of Claims & \numprint{2.61} \\
Avg. Number of Legal Provisions & \numprint{3.13} \\
Avg. Length in Judgment Document & \numprint{4243.46}  \\
\qquad in first-instance & \numprint{3672.48} \\
\qquad in second-instance & \numprint{4818.44} \\
\bottomrule
\end{tabular}
}
\caption{Dataset Statistics.}\label{tab:dataset}
\end{table}

\begin{figure}[t] 
    \centering
    \includegraphics[width=0.7\columnwidth]{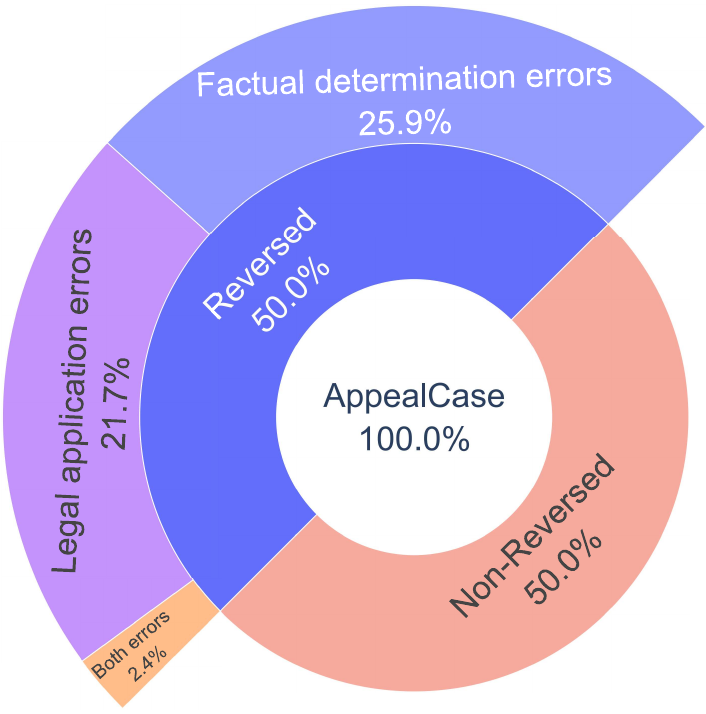}
    \caption{Distribution of reversal reasons in the AppealCase dataset.}
    \label{fig:statistic}
\end{figure}

The reasons for reversal can be divided into two categories: factual determination errors and legal application errors. Factual determination errors refer to situations where the original court made mistakes in identifying or understanding key or basic facts in the case, such as erroneous classifications, unclear findings, or insufficient evidentiary support. Specifically, this includes circumstances where the evidence on which the judgment is based is unreliable, insufficient, contradictory, or legally excluded, leading to deviations in determining the nature of the case or the allocation of rights and obligations between the parties \cite{Bodenheimer1987Jurisprudence}. Legal application errors include misjudgment of the nature of the law (e.g., misidentifying a tort dispute as a contract dispute), improper citation of legal provisions (e.g., applying a general law when a special law should be applied), or violations of the principle of non-retroactivity, all of which undermine the legal basis of the judgment.

The proportion of the top 10 causes of action in appeal cases is shown in Table \ref{tab:statistic_casecause}. Among them, the top cause, ``private lending disputes,'' accounts for 13.08\%, and the top ten causes together account for 60\%. The distribution of reasons for judgment reversal is shown in Figure \ref{fig:statistic}. Among these \numprint{10000} cases, 50\% were reversed, and there are 240 cases with reasons for reversal. Table \ref{tab:dataset} presents statistics of the dataset, showing that second-instance judgments are significantly longer than first-instance judgments, with an average of \numprint{4818} Chinese characters in second-instance judgments and an average of \numprint{3672} in first-instance judgments. Long documents pose a challenge to the context understanding ability of LegalAI models.

\begin{table*}[t]
\centering
\footnotesize
\renewcommand{\arraystretch}{1.0}
\scalebox{0.9}{
\begin{tabular}{l|ccc}
\toprule
\textbf{Task} & \textbf{Type} & \textbf{\# Sample} & \textbf{Metric} \\
\midrule
Judgment Reversal Prediction & Multi-label Classification & \numprint{10000} & Precision, Recall, F1-macro  \\
\qquad from the first-instance perspective & Multi-label Classification & \numprint{5481} & Precision, Recall, F1-macro  \\
\qquad from the second-instance perspective & Multi-label Classification & \numprint{4519} & Precision, Recall, F1-macro  \\
Legal Provision Recommendation & Multi-label Classification & \numprint{10000} & Precision, Recall, F1-samples \\
Legal Judgment Prediction & Single-label Classification & \numprint{26143} & Accuracy, Precision, Recall, F1-macro \\
Court View Generation & Text Generation & \numprint{10000} & ROUGE-\{1, 2, L\}, BLEU-\{1, 2, 3\}, LLM Judger\\
\bottomrule
\end{tabular}
}
\caption{Task list of AppealCase, including three types of tasks: multi-label classification, single-label classification, and text generation. }\label{tab:tasks}
\end{table*}

\subsection{Dataset Construction}

In the process of constructing AppealCase, we structured the documents, matched first-instance and second-instance judgments based on case numbers, and annotated the documents using a multi-layer legal annotation framework.

\subsubsection{Document Structuring and Matching}

We divided the judgment documents into six sections—header, claim, fact description, court's view, judgment, and footer—based on keyword patterns, and discarded documents that could not be segmented accordingly.

Then, we used regular expressions to extract the unique first-instance case number mentioned in second-instance documents and retrieved the corresponding first-instance documents to establish case pairs.

\subsubsection{Label Annotation}

We provided five types of annotations for each case in AppealCase through a multi-layer annotation framework consisting of rules/regular expressions, large language models, and expert annotation.

\paragraph{Judgment Reversal}

The label ``Judgment Reversal'' reflects whether the second-instance court has revoked the first-instance judgment.

We annotate whether the second-instance judgment has been reversed based on rules extracted from the judgment documents. If the judgment paragraph contains the keywords ``appeal dismissed'' and ``original judgment upheld,'' it is labeled as not reversed, otherwise, it is labeled as reversed.

\paragraph{Reasons for Reversal}

For cases that have been reversed, the label ``Reasons for Reversal'' reflects the reasons for the reversal, and there may be multiple reasons.

For these documents, we extract the reasons for reversal from the court’s view section according to rules. If the rules do not match, we use an LLM for annotation. The LLM generates ten responses through ten rounds of sampling. If all responses are identical, the result is directly adopted. If there is low consistency among the responses, the annotation is referred to manual review.

\paragraph{Claims}

The label “Claims” annotates each claim in the first-instance documents, indicating whether it was supported by the first-instance court and whether it was supported by the second-instance court.

The result is one of three types: fully supported, partially supported, or not supported. We first use regular expressions to split each claim in the first instance. Then, we use an LLM to annotate whether each claim was supported in the first-instance court and the second-instance court. The LLM generates ten responses through ten rounds of sampling. If all responses are identical, the result is directly adopted. If there is low consistency among the responses, the annotation is referred to manual review.

\paragraph{Legal Provisions} 

The label ``Legal Provisions'' annotates the legal articles cited in the second-instance judgment documents.

We use regular expressions to extract the legal provisions cited in the court's view and judgment sections of the second-instance judgment documents, and separately annotate the name of the law and the article number cited.

\paragraph{New Information}

The label ``New Information'' indicates whether new information was submitted during the appeal process, which is usually new evidence. The new information may affect the second-instance court's determination of the facts, or it may be excluded.

We first annotate whether new information was introduced based on rules. For cases where annotation cannot be performed by rules, we use LLM for annotation. Similar to other labels, samples with low consistency are manually annotated by legal experts.

\subsection{Human Expert Evaluation}

We invited three legal experts, including law graduate students and professional lawyers, to review 500 cases randomly sampled from the AppealCase dataset. The experts assessed the completeness and consistency of the first-instance and second-instance documents for each case, as well as the accuracy of the five annotations. Over 99\% of the sampled cases were marked as accurate by the experts, with incorrect samples being corrected accordingly.

\section{Task Definition}

To better apply the AppealCase dataset in the legal domain, we propose five new LegalAI tasks, as shown in Table \ref{tab:tasks}, including: Judgment Reversal Prediction from the perspective of the first-instance and the second-instance, Legal Provision Recommendation, Legal Judgment Prediction, and Court View Generation in the second-instance scenario. We provide task examples in Appendix \ref{appendix:task}.

\subsection{Judgment Reversal Prediction}

The judgment reversal prediction task helps reduce the first-instance misjudgment rate and assists in adjudication in the second-instance. To reflect the differences in information acquisition during actual trials, we divide the data based on whether new information is introduced at the second-instance stage: cases with new information are regarded as the second-instance perspective, while those without are regarded as the first-instance perspective, reflecting the differences in information available to the courts at the two trial levels.

\begin{problem} [Judgment Reversal Prediction from the first-instance perspective]
Given the first-instance document and the second-instance claim, the task is to predict the reasons for reversal.
\end{problem}

\begin{problem} [Judgment Reversal Prediction from the second-instance perspective]
Given the first-instance document and the second-instance claim and fact description, which contains new information introduced in the second instance, the task is to predict the reasons for reversal.
\end{problem}

\subsection{Legal Provision Recommendation}

Legal provisions form the foundation of judicial decisions. In the legal provision prediction task, our goal is to accurately select the most relevant legal provisions based on the facts of the case. To achieve this, we prepare 10 candidate options for each case, with the correct option coming from the legal provisions annotated for the case, while the remaining distractors are randomly selected from legal provisions involved in other cases.

\begin{problem} [Legal Provision Recommendation]
Given the candidate legal provisions, and the second-instance header, claim, and fact description, the task is to select the relevant legal provisions.
\end{problem}

\begin{table*}[t]
\centering
\footnotesize
\renewcommand{\arraystretch}{1.00}
\scalebox{0.85}{
\begin{tabular}{c|l|ccc|ccc}
\hline
\multirow{2}{*}{\textbf{Category}} & \multicolumn{1}{c|}{\multirow{2}{*}{\textbf{Model}}}  & \multicolumn{3}{c|}{\textbf{First-instance Perspective}} & \multicolumn{3}{c}{\textbf{Second-instance Perspective}} \\
 & \multicolumn{1}{c|}{} & \multicolumn{1}{c}{\textbf{Precision}} & \multicolumn{1}{c}{\textbf{Recall}} & \multicolumn{1}{c|}{\textbf{F1}} & \multicolumn{1}{c}{\textbf{Precision}} & \multicolumn{1}{c}{\textbf{Recall}} & \multicolumn{1}{c}{\textbf{F1}} \\ \hline
\multirow{9}{*}{
\makecell[c]{Non- \\ Reasoning}} & DeepSeek-V3 & 44.94 & 41.87 & 42.53 & 55.56 & 54.97 & 54.49\\
 & Qwen2.5-72B & 47.62 & 41.23 & 40.49 & 55.84 & \underline{59.45} & \textbf{\underline{57.40}} \\
 & LLaMA3.3-70B & 40.75 & 45.22 & 34.85 & 50.42 & 56.94 & 48.08\\ \cline{2-8} 
 & GPT-4.1 & \textbf{\underline{51.93}} & 36.53 & 32.58 & \textbf{\underline{60.28}} & 47.30 & 44.80 \\
 & GLM-4-Air & 38.64 & 42.09 & 33.24 & 42.08 & 41.10 & 38.72 \\
 & Doubao-1-5-pro & 42.57 & \underline{47.42} & \textbf{\underline{44.57}} & 55.28 & 59.27 & 54.73 \\ \cline{2-8} 
 & Baichuan2-7B & 26.00 & 33.37 & 26.60 & 34.99 & 34.87 & 25.90 \\
 & Qwen2.5-7B & 38.28 & 34.31 & 30.42 & 46.02 & 41.56 & 40.18 \\
 & Llama3.1-8B & 34.38 & 34.34 & 18.43 & 41.45 & 36.06 & 28.05 \\ \hline
\multirow{9}{*}{Reasoning} & DeepSeek-R1 & \underline{44.17} & 43.54 & \underline{43.06} & 54.03 & 55.56 & \underline{54.77} \\
 & R1-Distill-Qwen-32B & 40.01 & 45.61 & 40.58 & 49.28 & 57.36 & 52.07\\
 & QwQ-32B & 42.31 & \textbf{\underline{51.41}} & 40.30 & 52.38 & 54.79 & 49.95 \\
 & Qwen3-32B & 40.06 & 49.19 & 39.30 & 49.91 & 59.19 & 50.64\\ \cline{2-8} 
 & GLM-Z1-Air & 39.34 & 43.34 & 39.46 & 49.98 & 54.34 & 48.90 \\
 & GPT-o4-mini & 43.67 & 40.49 & 40.36 & \underline{54.31} & 49.16 & 47.85 \\
 & Grok-3-mini & 37.23 & 49.37 & 37.41 & 48.97 & \textbf{\underline{62.64}} & 53.69\\ \cline{2-8} 
 & R1-Distill-Qwen-7B & 35.82 & 44.84 & 37.71 & 37.87 & 52.98 & 41.68 \\
 & Qwen3-8B & 39.79 & 51.34 & 36.03 & 49.21 & 55.44 & 47.35 \\ \hline
\multirow{2}{*}{Domain} & DISC-LawLLM & 32.51 & 33.85 & 30.05 & \underline{35.11} & \underline{34.11} & \underline{23.09} \\
 & Wisdom Interrogatory & \underline{32.92} & \underline{34.20} & \underline{30.47} & 33.74 & 33.93 & 22.52 \\ \hline
\end{tabular}
}
\caption{Performance on judgment reversal prediction from the first-instance and second-instance perspective. Overall best and best results for each category are in \textbf{bold} and \underline{underline}, respectively.}\label{tab:JRP}
\end{table*}

\subsection{Legal Judgment Prediction}

Unlike previous judgment prediction tasks \cite{cui2023survey}, the task here focuses on adjudication results of each claim from the first-instance at the second-instance stage. We examine whether each claim continues to be supported in the second instance, with results categorized as fully, partially, or not supported. This provides a new perspective for in-depth analyzing changes in claim support across different trial levels.

\begin{problem} [Legal Judgment Prediction]
Given the first-instance fact description, the second-instance claim and fact description, and the claim to be judged, the task is to predict whether this claim is supported in the second instance.
\end{problem}

\subsection{Court View Generation}

Unlike the first-instance court view generation task \cite{wu-etal-2020-de}, second-instance court view generation not only requires the independent application of law, but also a review of the fact-finding and legal application in the first-instance judgment, and, when necessary, clarification of the specific reasons for reversal. The second-instance court view generation task requires the LegalAI model to reflect the unique perspective and logic of second-instance review and re-judgment, which is of great significance for understanding the supervisory and remedial mechanisms of the judicial trial system.

\begin{problem} [Court View Generation]
Given the second-instance claim and fact description, the task is to generate the second-instance court’s view, reasons for reversal, and judgment. The summary of the first-instance document is usually already included in the second-instance fact description.
\end{problem}

\section{Experiments}

We conducted a comprehensive evaluation of the five new LegalAI tasks across 20 models. The experimental details can be found in Appendix \ref{appendix:experiment}.

\subsection{Model Categories and Evaluation Setup}

We selected a diverse set of pretrained language models as baselines, classified into three categories: non-reasoning models, reasoning models, and domain-specific models, covering representative architectures at different scales. Detailed model providers, model versions, and references can be found in Appendix  \ref{appendix:models}.

\paragraph{Non-Reasoning Models} 

The open-source models include DeepSeek-V3, Qwen2.5-72B, and LLaMA3.3-70B, while the closed-source models comprise GPT-4.1, GLM-4-Air, and Doubao-1.5-pro. To support low-resource environments, we also introduce three small open-source models: Baichuan2-7B, Qwen2.5-7B, and LLaMA3.1-8B.

\paragraph{Reasoning Models}

These models possess advanced reasoning capabilities and are designed for complex inference tasks. Open-source models include DeepSeek-R1, R1-Distill-Qwen-32B, QwQ-32B, and Qwen3-32B, while closed-source reasoning models include GLM-Z1-Air, GPT-o4-mini, and Grok-3-mini. Smaller-scale LLMs in this category include R1-Distill-Qwen-7B and Qwen3-8B.

\paragraph{Domain-Specific Models}

These models are pretrained and fine-tuned on large-scale legal corpora, including DISC-LawLLM \cite{yue2023disc} and Wisdom Interrogatory \cite{wisdom2024modelcard}.

\begin{table}[t]
\centering
\footnotesize
\renewcommand{\arraystretch}{1.05}
\scalebox{0.88}{
\begin{tabular}{cc|cc}
\hline
\textbf{Tasks} & \textbf{Metrics} & \textbf{BERT} & \textbf{Qwen3} \\ \hline
\multirow{3}{*}{\makecell[c]{First- \\ instance \\ Perspective}} & Precision & \textbf{58.24} & 56.27 \\
 & Recall & 49.56 & \textbf{56.93} \\
 & F1 & 52.70 & \textbf{56.10} \\ \hline
\multirow{3}{*}{\makecell[c]{Second- \\ instance \\ Perspective}} & Precision & 60.03 & \textbf{67.97} \\
 & Recall & 55.74 & \textbf{63.00} \\
 & F1 & 56.20 & \textbf{63.46} \\
\hline
\end{tabular}
}
\caption{Performance of fine-tuned models on judgment reversal prediction.}\label{tab:finetune}
\end{table}

\begin{table*}[t]
\centering
\footnotesize
\renewcommand{\arraystretch}{1.00}
\scalebox{0.80}{
\begin{tabular}{c|l|ccc|cccc}
\hline
\multirow{2}{*}{\textbf{Category}} & \multicolumn{1}{c|}{\multirow{2}{*}{\textbf{Model}}}  & \multicolumn{3}{c|}{\textbf{Provision Recommendation}} & \multicolumn{4}{c}{\textbf{Judgment Prediction}} \\
 & \multicolumn{1}{c|}{} &  \multicolumn{1}{c}{\textbf{Precision}} & \multicolumn{1}{c}{\textbf{Recall}} & \multicolumn{1}{c|}{\textbf{F1}} & \multicolumn{1}{c}{\textbf{Accuracy}} & \multicolumn{1}{c}{\textbf{Precision}} & \multicolumn{1}{c}{\textbf{Recall}} & \multicolumn{1}{c}{\textbf{F1}} \\ \hline
\multirow{9}{*}{\makecell[c]{Non- \\ Reasoning}} & DeepSeek-V3 &  87.22 & 78.00 & 77.88 & 64.26 & 69.08 & 63.00 & 61.91 \\
 & Qwen2.5-72B &  73.84 & 82.31 & 72.24 & 66.67 & 67.19 & 66.39 & 66.33 \\
 & LLaMA3.3-70B &  59.50 & 46.15 & 45.21 & 58.59 & 59.38 & 58.67 & 58.24 \\ \cline{2-9} 
 & GPT-4.1 &  73.18 & 69.97 & 65.59 & 68.19 & 68.35 & 68.49 & 68.19 \\
 & GLM-4-Air &  67.96 & 58.02 & 55.61 & 46.50 & 52.07 & 47.97 & 44.19\\
 & Doubao-1-5-pro &  \underline{\textbf{90.50}} & \underline{\textbf{83.00}} & \underline{\textbf{83.65}} & \underline{\textbf{70.93}} & \underline{\textbf{71.49}} & \underline{\textbf{70.34}} & \underline{\textbf{70.62}} \\ \cline{2-9} 
 & Baichuan2-7B &  48.22 & 24.77 & 26.90 & 43.41 & 42.64 & 41.38 & 38.25 \\
 & Qwen2.5-7B &  84.32 & 67.35 & 69.44 & 53.34 & 59.08 & 52.48 & 48.79 \\
 & Llama3.1-8B &  49.21 & 34.80 & 35.31 & 42.84 & 48.38 & 39.56 & 36.23 \\ \hline
\multirow{9}{*}{Reasoning} & DeepSeek-R1 &  \underline{84.85} & 60.66 & 66.15 & 63.33 & 65.95 & 63.65 & 62.41 \\
 & R1-Distill-Qwen-32B &  76.09 & 63.08 & 63.41 & 62.35 & 63.34 & 63.49 & 62.31 \\
 & QwQ-32B &  64.97 & \underline{83.93} & 64.05 & 56.57 & 59.69 & 58.35 & 56.16 \\
 & Qwen3-32B &  73.92 & 75.68 & \underline{68.46} & 62.81 & 62.95 & 63.31 & 62.77 \\ \cline{2-9} 
 & GLM-Z1-Air &  72.62 & 52.02 & 56.31 & 57.16 & 60.98 & 58.33 &  56.75\\
 & GPT-o4-mini &  60.81 & 47.86 & 49.30 & 54.13 & 60.17 & 56.51 & 53.01 \\
 & Grok-3-mini &  59.75 & 63.99 & 56.02 & \underline{66.20} & \underline{66.83} & \underline{66.07} & \underline{65.90} \\ \cline{2-9} 
 & R1-Distill-Qwen-7B &  44.92 & 31.45 & 31.16 & 36.35 & 42.41 & 39.17 & 32.71 \\
 & Qwen3-8B &  57.68 & 62.44 & 50.06 & 59.86 & 60.41 & 60.30 & 59.73 \\ \hline
\multirow{2}{*}{Domain} & DISC-LawLLM &  22.72 & 58.56 & 27.24 & \underline{34.95} & \underline{37.45} & \underline{37.98} & \underline{29.93} \\
 & Wisdom Interrogatory &  \underline{23.01} & \underline{58.91} & \underline{27.31} & 34.75 & 36.15 & 37.79 & 29.64 \\ \hline
\end{tabular}
}
\caption{Performance on legal provision recommendation and legal judgment prediction.}\label{tab:LPR}
\end{table*}

\begin{table*}[t]
\centering
\footnotesize
\renewcommand{\arraystretch}{1.00}
\scalebox{0.76}{
\begin{tabular}{c|l|ccccccc}
\hline
\multirow{2}{*}{\textbf{Category}} & \multicolumn{1}{c|}{\multirow{2}{*}{\textbf{Model}}}  & \multicolumn{7}{c}{\textbf{Court View Generation}} \\
 & \multicolumn{1}{c|}{} & \multicolumn{1}{c}{\textbf{ROUGE-1}} & \multicolumn{1}{c}{\textbf{ROUGE-2}} & \multicolumn{1}{c}{\textbf{ROUGE-L}} & \multicolumn{1}{c}{\textbf{BLEU-1}} & \multicolumn{1}{c}{\textbf{BLEU-2}} & \multicolumn{1}{c}{\textbf{BLEU-3}} & \multicolumn{1}{c}{\textbf{LLM}} \\ \hline
\multirow{9}{*}{\makecell[c]{Non- \\ Reasoning}} & DeepSeek-V3 & 50.43 & 24.41 & 29.50 & 37.23 & 26.01 & 19.65 & 7.73\\
 & Qwen2.5-72B &  \underline{\textbf{52.66}} & \underline{\textbf{27.61}} & \underline{\textbf{32.21}} & 40.14 & \underline{\textbf{29.14}} & \underline{\textbf{22.78}} & 7.67 \\
 & LLaMA3.3-70B &  41.33 & 18.30 & 23.68 & 23.04 & 15.33 & 11.16 & 7.09 \\ \cline{2-9} 
 & GPT-4.1 &  47.40 & 19.22 & 24.38 & 38.65 & 24.80 & 17.31 & \underline{7.85} \\
 & GLM-4-Air &  41.42 & 16.85 & 22.41 & 23.97 & 15.27 & 10.59 & 7.04 \\
 & Doubao-1-5-pro &  50.74 & 25.26 & 29.81 & \underline{\textbf{40.93}} & 28.71 & 21.84 & 7.78 \\ \cline{2-9} 
 & Baichuan2-7B &  43.70 & 22.14 & 26.99 & 26.71 & 19.21 & 15.07 & 7.18 \\
 & Qwen2.5-7B &  49.26 & 24.84 & 29.65 & 34.59 & 24.70 & 19.08 & 7.55 \\
 & Llama3.1-8B &  31.66 & 12.44 & 17.11 & 13.62 & 8.62 & 6.09 & 5.90 \\ \hline
\multirow{9}{*}{Reasoning} & DeepSeek-R1 &  46.98 & 20.60 & 26.40 & 34.62 & 22.94 & 16.62 & 7.79 \\
 & R1-Distill-Qwen-32B &  46.17 & \underline{22.38} & \underline{27.56} & 27.49 & 19.39 & 14.65 & 7.82 \\
 & QwQ-32B &  \underline{48.32} & 21.99 & 27.55 & 35.75 & \underline{24.16} & \underline{17.66} & \textbf{\underline{7.96}} \\
 & Qwen3-32B &  47.24 & 20.44 & 25.70 & \underline{36.32} & 23.82 & 17.11 & 7.91 \\ \cline{2-9} 
 & GLM-Z1-Air &  44.50 & 18.38 & 23.75 & 33.03 & 21.15 & 15.13 & 7.89 \\
 & GPT-o4-mini &  42.18 & 14.94 & 20.93 & 29.41 & 17.52 & 11.60 & 7.68 \\
 & Grok-3-mini &  47.44 & 20.11 & 24.92 & 35.11 & 22.98 & 16.33 & 7.70 \\ \cline{2-9} 
 & R1-Distill-Qwen-7B &  33.78 & 12.11 & 18.29 & 15.24 & 9.05 & 5.87 & 7.11 \\
 & Qwen3-8B &  46.55 & 20.05 & 25.19 & 34.85 & 22.69 & 16.07 & 7.88 \\ \hline
\multirow{2}{*}{Domain} & DISC-LawLLM &  \underline{29.59} & \underline{11.25} & 15.99 & 11.97 & 7.38 & 5.16 & 6.11 \\
 & Wisdom Interrogatory &  29.52 & 11.20 & \underline{15.91} & \underline{12.02} & \underline{7.41} & \underline{5.18} & \underline{6.12} \\ \hline
\end{tabular}
}
\caption{Performance on court view generation.}\label{tab:CVG}
\end{table*} 

\subsection{Results on Judgment Reversal Prediction}

From Table \ref{tab:JRP}, we observe the following results: \textbf{1)} In the judgment reversal prediction task, all models perform poorly, highlighting the challenge of this task. Under the first-instance perspective, all models have F1 scores below 50\%. Under the second-instance perspective, more than half of the models have F1 scores below 50\%, with the highest only reaching 57.40\%. \textbf{2)} Existing domain-specific models are limited by their context window length and cannot process long judicial documents. \textbf{3)} The models perform better under the second-instance perspective than under the first-instance perspective, which may be due to the second-instance fact description containing more information, especially the summary of the first-instance document.

We also fine-tuned two small-scale models, bert-base-chinese \cite{devlin-etal-2019-bert} and Qwen3-0.6b \cite{yang2025qwen3}, using 80\% of the data. As shown in Table \ref{tab:finetune}, the experiments indicate that fine-tuning can improve model performance on judgment reversal prediction, but the improvement remains limited, with the average F1 score still below 60\%.

\subsection{Results on Provision Recommendation and Judgment Prediction}

From Table \ref{tab:LPR}, we can conclude that: \textbf{1)} Doubao-1.5-pro performs well on both legal provision recommendation and legal judgment prediction, with F1 scores reaching 83.65\% and 70.62\%, respectively. \textbf{2)} Since the second-instance fact description usually includes a summary of the first-instance judgment (including the first-instance claim, cited legal provisions, and judgment result), the model's performance on these two tasks is relatively high. However, more than half of the models still have F1 scores below 60\%.

\subsection{Results on Court View Generation}

From Table \ref{tab:CVG}, we find that: \textbf{1)} In court view generation, the SOTA model Qwen2.5-72B achieves average ROUGE and BLEU scores of less than 40\%, indicating that the court views generated by LLMs still have significant gaps in structure and wording compared to real judicial documents. \textbf{2)} Models whose training data focuses on Chinese corpora have an advantage in generation tasks compared to those focusing on other languages.

\begin{figure*}[t] 
    \centering
    \includegraphics[width=1.0\textwidth]{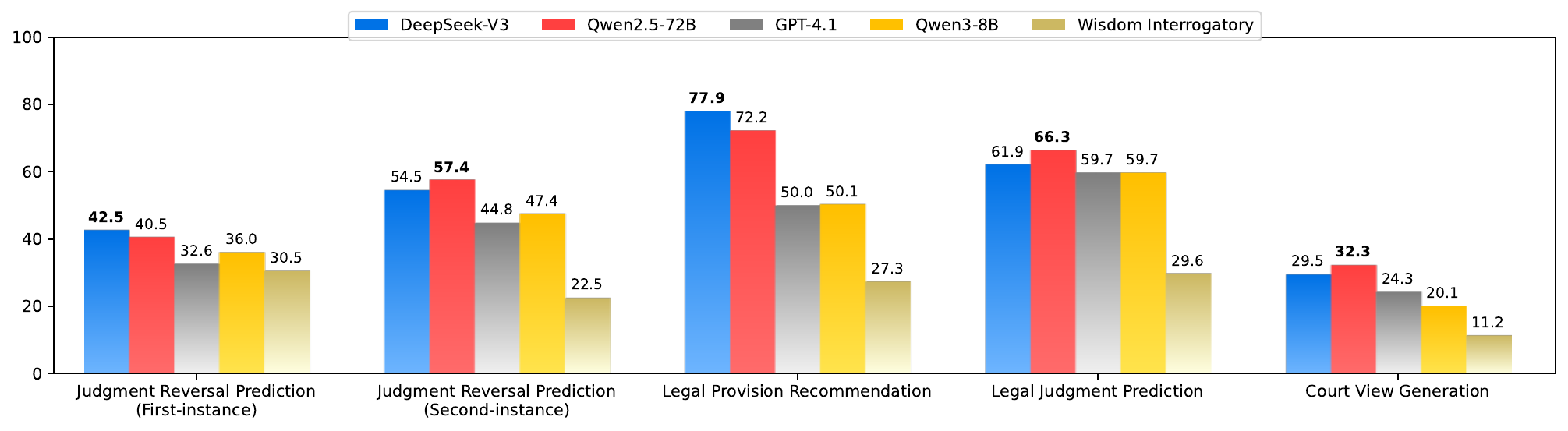}
    \caption{Performance of different models on the five tasks.}
    \label{fig:model}
\end{figure*}

\subsection{Comparison between tasks and models}

We compared the F1 and ROUGE-L scores of several models across five tasks, as shown in Figure \ref{fig:model}. The results indicate that: \textbf{1)} large-scale models outperform small-scale models. \textbf{2)} Models with a higher proportion of Chinese training data outperform others. \textbf{3)} The performance of domain-specific models is limited by context length.

Among the five tasks, judgment reversal prediction best demonstrates the model’s ability to capture the characteristics of appellate scenarios and is also the most challenging. Since first-instance judgment documents contain references to legal provisions and initial judgment results, they provide useful information for the models. As a result, performance on the legal provision recommendation and legal judgment prediction tasks is relatively higher, though these tasks remain challenging. The relatively low textual similarity in the court view generation task indicates that current LLMs still face challenges in generating judicial documents with appropriate structure and wording.

\section{Related Work}

\subsection{Legal Artificial Intelligence Research}

Recent advances in large-scale pre-trained models like BERT and GPT have boosted LegalAI tasks such as provision matching, fact extraction, and judgment prediction \cite{zhong2020does}. Specialized models for long legal texts \cite{xiao2021lawformer}, integrated retrieval-judgment systems \cite{qin2024explicitly}, and knowledge-enhanced prompting for Chinese cases \cite{sun2024chinese} have been proposed. Graph-based approaches incorporating domain knowledge, e.g., GraphWordBag for confusing charge prediction \cite{li2024graphwordbagintroducing} and the constraint-enhanced GJudge model \cite{tong2024legal}, improve prediction accuracy. Causal inference methods enhance consistency and interpretability in European Court of Human Rights cases \cite{santosh2022deconfounding}. Despite comprehensive surveys summarizing these advances and challenges \cite{feng2022legal}, research on complex appellate tasks like second-instance retrieval and court view generation remains limited, as emphasized by CAIL2024 \footnote{\url{http://cail.cipsc.org.cn/task_summit.html?raceID=3&cail_tag=2024}}.

\subsection{Legal AI Benchmarks and Datasets}

Current LegalAI benchmarks primarily focus on independent single-stage tasks, with variations in coverage and evaluation depth. Internationally, LegalBench \cite{guha2023legalbench} based on U.S. federal law and EURLEX \cite{chalkidis-etal-2020-legal} centered on EU legislation provide extensive classification annotations, serving as foundational resources. Domestically, the annual CAIL evaluation covers a wide range of tasks and serves as the main data source for LawBench \cite{fei2024lawbench}. LexEval \cite{li2024lexeval} and LegalAgentBench \cite{li2024legalagentbench} focus on Chinese law but are limited to first-instance judgments, lacking analysis of second-instance reasoning and cross-instance joint evaluation. Existing surveys \cite{JunyunCui2022ASO} systematically summarize datasets, metrics, models, and challenges in legal judgment prediction, providing key references for LegalAI benchmark and dataset research.

\section{Conclusion}

This paper addresses the appellate process, which is generally neglected in the current LegalAI field, by constructing the realistically paired AppealCase dataset. The dataset covers \numprint{10000} pairs of first-instance and second-instance judgments across 91 categories of civil cases, and, considering the uniqueness of the appellate scenario, proposes a detailed labeling schema and conducts annotation. On this basis, we propose five novel LegalAI tasks specifically designed for the appeal context and conduct systematic evaluations on 20 mainstream models. Experimental results show that existing LLMs still have considerable room for improvement on key tasks such as judgment reversal prediction, highlighting the complexity and challenge of the appellate scenario. We hope that the AppealCase dataset will promote in-depth research in LegalAI for appellate case analysis and contribute to the consistency and fairness of judicial decisions. In the future, as model capabilities continue to improve, research based on this dataset is expected to provide more powerful intelligent assistance to actual judicial practice.
\section*{Limitations}

The limitations of AppealCase are summarized as follows:

\begin{itemize}[itemsep=3pt,topsep=3pt,parsep=0pt]
\item The proposed AppealCase is currently focused on the laws and cases of the People’s Republic of China and has not yet been extended to the judicial systems of other countries or regions. Considering that China is a typical representative of the civil law system, and that the appeal system has structural similarities in both civil law and common law systems, this research thus holds a certain degree of cross-legal-system representativeness.

\item AppealCase contains a selected set of 91 civil causes of action and does not cover all civil causes of action, nor does it include criminal, administrative, or other types of cases. Although the selected causes of action account for the vast majority of causes in civil litigation, they may not generalize to other causes of action. Future work will explore extending the annotation system and data to a wider range of causes of action.

\end{itemize}
\section*{Ethics Statement}

\paragraph{Data Privacy} All data used in this study are publicly available sourced from the China Judgments Online platform. In accordance with the regulations of the Supreme People's Court of the People's Republic of China regarding the publication of court judgments on the internet, information such as individuals' home addresses, contact information, identification numbers, bank account numbers, trade secrets, and information related to minors has been removed. This study does not involve any user-related data or private data.

\paragraph{Intended Use}

The appeal system is an important safeguard for judicial fairness. The AppealCase dataset proposed in this study is centered on the pairing of first-instance and second-instance court judgments, aiming to support academic research in the areas of appeal analysis and judicial consistency. It provides data and benchmarks for LegalAI tasks in second-instance scenarios such as reversal prediction, legal opinion generation and judgment prediction. The dataset can be used to explore the evolution of case facts, points of contention, and court opinions across different court levels, assisting AI models in better understanding the legal system. We remind researchers that any malicious use or abuse of this dataset should be prohibited.

\bibliography{custom}

\begin{thebibliography}{43}
\expandafter\ifx\csname natexlab\endcsname\relax\def\natexlab#1{#1}\fi

\bibitem[{Bodenheimer et~al.(1987)Bodenheimer, Deng, and Ji}]{Bodenheimer1987Jurisprudence}
E.~Bodenheimer, Zhenlai Deng, and Jingwu Ji. 1987.
\newblock \emph{Jurisprudence: Philosophy of Law and Its Methods}.
\newblock Huaxia Publishing House.

\bibitem[{B{\"u}ttner and Habernal(2024)}]{buttner-habernal-2024-answering}
Marius B{\"u}ttner and Ivan Habernal. 2024.
\newblock \href {https://aclanthology.org/2024.eacl-long.122/} {Answering legal questions from laymen in {G}erman civil law system}.
\newblock In \emph{Proceedings of the 18th Conference of the European Chapter of the Association for Computational Linguistics (Volume 1: Long Papers)}, pages 2015--2027, St. Julian{'}s, Malta. Association for Computational Linguistics.

\bibitem[{Chalkidis et~al.(2020)Chalkidis, Fergadiotis, Malakasiotis, Aletras, and Androutsopoulos}]{chalkidis-etal-2020-legal}
Ilias Chalkidis, Manos Fergadiotis, Prodromos Malakasiotis, Nikolaos Aletras, and Ion Androutsopoulos. 2020.
\newblock \href {https://doi.org/10.18653/v1/2020.findings-emnlp.261} {{LEGAL}-{BERT}: The muppets straight out of law school}.
\newblock In \emph{Findings of the Association for Computational Linguistics: EMNLP 2020}, pages 2898--2904, Online. Association for Computational Linguistics.

\bibitem[{Cui et~al.(2022)Cui, Shen, Nie, Wang, Wang, and Chen}]{JunyunCui2022ASO}
Junyun Cui, Xiaoyu Shen, Feiping Nie, Zheng Wang, Jinglong Wang, and Yulong Chen. 2022.
\newblock \href {https://doi.org/10.48550/arXiv.2204.04859} {A survey on legal judgment prediction: Datasets, metrics, models and challenges}.
\newblock \emph{CoRR}, abs/2204.04859.

\bibitem[{Cui et~al.(2023)Cui, Shen, and Wen}]{cui2023survey}
Junyun Cui, Xiaoyu Shen, and Shaochun Wen. 2023.
\newblock A survey on legal judgment prediction: Datasets, metrics, models and challenges.
\newblock \emph{IEEE Access}, 11:102050--102071.

\bibitem[{Devlin et~al.(2019)Devlin, Chang, Lee, and Toutanova}]{devlin-etal-2019-bert}
Jacob Devlin, Ming-Wei Chang, Kenton Lee, and Kristina Toutanova. 2019.
\newblock \href {https://doi.org/10.18653/v1/N19-1423} {{BERT}: Pre-training of deep bidirectional transformers for language understanding}.
\newblock In \emph{Proceedings of the 2019 Conference of the North {A}merican Chapter of the Association for Computational Linguistics: Human Language Technologies, Volume 1 (Long and Short Papers)}, pages 4171--4186, Minneapolis, Minnesota. Association for Computational Linguistics.

\bibitem[{Fei et~al.(2024)Fei, Shen, Zhu, Zhou, Han, Huang, Zhang, Chen, Yin, Shen et~al.}]{fei2024lawbench}
Zhiwei Fei, Xiaoyu Shen, Dawei Zhu, Fengzhe Zhou, Zhuo Han, Alan Huang, Songyang Zhang, Kai Chen, Zhixin Yin, Zongwen Shen, et~al. 2024.
\newblock Lawbench: Benchmarking legal knowledge of large language models.
\newblock In \emph{Proceedings of the 2024 Conference on Empirical Methods in Natural Language Processing}, pages 7933--7962.

\bibitem[{Feng et~al.(2022)Feng, Li, and Ng}]{feng2022legal}
Yi~Feng, Chuanyi Li, and Vincent Ng. 2022.
\newblock Legal judgment prediction: A survey of the state of the art.
\newblock In \emph{Proceedings of the Thirty-First International Joint Conference on Artificial Intelligence, IJCAI-22}, pages 5461--5469.

\bibitem[{GLM et~al.(2024)GLM, Zeng, Xu, Wang, Zhang, Yin, Zhang, Rojas, Feng, Zhao et~al.}]{glm2024chatglm}
Team GLM, Aohan Zeng, Bin Xu, Bowen Wang, Chenhui Zhang, Da~Yin, Dan Zhang, Diego Rojas, Guanyu Feng, Hanlin Zhao, et~al. 2024.
\newblock Chatglm: A family of large language models from glm-130b to glm-4 all tools.
\newblock \emph{arXiv preprint arXiv:2406.12793}.

\bibitem[{Guha et~al.(2023)Guha, Nyarko, Ho, R{\'e}, Chilton, Chohlas-Wood, Peters, Waldon, Rockmore, Zambrano et~al.}]{guha2023legalbench}
Neel Guha, Julian Nyarko, Daniel Ho, Christopher R{\'e}, Adam Chilton, Alex Chohlas-Wood, Austin Peters, Brandon Waldon, Daniel Rockmore, Diego Zambrano, et~al. 2023.
\newblock Legalbench: A collaboratively built benchmark for measuring legal reasoning in large language models.
\newblock \emph{Advances in Neural Information Processing Systems}, 36:44123--44279.

\bibitem[{Guo et~al.(2025)Guo, Yang, Zhang, Song, Zhang, Xu, Zhu, Ma, Wang, Bi et~al.}]{guo2025deepseek}
Daya Guo, Dejian Yang, Haowei Zhang, Junxiao Song, Ruoyu Zhang, Runxin Xu, Qihao Zhu, Shirong Ma, Peiyi Wang, Xiao Bi, et~al. 2025.
\newblock Deepseek-r1: Incentivizing reasoning capability in llms via reinforcement learning.
\newblock \emph{arXiv preprint arXiv:2501.12948}.

\bibitem[{Jiang and Shao(1998)}]{Jiang1998ResearchAchievements}
Wei Jiang and Ming Shao. 1998.
\newblock Achievements and issues in the study of civil procedure law and its development.
\newblock \emph{China Law Review}, (4):18--22.

\bibitem[{Lax(2007)}]{lax2007constructing}
Jeffrey~R Lax. 2007.
\newblock Constructing legal rules on appellate courts.
\newblock \emph{American Political Science Review}, 101(3):591--604.

\bibitem[{Li et~al.(2024{\natexlab{a}})Li, Chen, Wu, Cai, Zhou, Wu, and Kuang}]{li2024graphwordbagintroducing}
Ang Li, Qiangchao Chen, Yiquan Wu, Ming Cai, Xiang Zhou, Fei Wu, and Kun Kuang. 2024{\natexlab{a}}.
\newblock \href {http://arxiv.org/abs/2403.04369} {From graph to word bag: Introducing domain knowledge to confusing charge prediction}.

\bibitem[{Li et~al.(2024{\natexlab{b}})Li, Wu, Liu, Kuang, Wu, and Cai}]{li-etal-2024-enhancing}
Ang Li, Yiquan Wu, Yifei Liu, Kun Kuang, Fei Wu, and Ming Cai. 2024{\natexlab{b}}.
\newblock \href {https://aclanthology.org/2024.lrec-main.522/} {Enhancing court view generation with knowledge injection and guidance}.
\newblock In \emph{Proceedings of the 2024 Joint International Conference on Computational Linguistics, Language Resources and Evaluation (LREC-COLING 2024)}, pages 5896--5906, Torino, Italia. ELRA and ICCL.

\bibitem[{Li et~al.(2024{\natexlab{c}})Li, Chen, Yang, Ai, Jia, Liu, Lin, Wu, Yuan, Hu et~al.}]{li2024legalagentbench}
Haitao Li, Junjie Chen, Jingli Yang, Qingyao Ai, Wei Jia, Youfeng Liu, Kai Lin, Yueyue Wu, Guozhi Yuan, Yiran Hu, et~al. 2024{\natexlab{c}}.
\newblock Legalagentbench: Evaluating llm agents in legal domain.
\newblock \emph{arXiv preprint arXiv:2412.17259}.

\bibitem[{Li et~al.(2024{\natexlab{d}})Li, Chen, Ai, Wu, Zhang, and Liu}]{li2024lexeval}
Haitao Li, You Chen, Qingyao Ai, Yueyue Wu, Ruizhe Zhang, and Yiqun Liu. 2024{\natexlab{d}}.
\newblock Lexeval: A comprehensive chinese legal benchmark for evaluating large language models.
\newblock In \emph{Proceedings of the Thirty-eighth Conference on Neural Information Processing Systems (NeurIPS 2024), Datasets and Benchmarks Track}.

\bibitem[{Li(2008)}]{Li2008BurdenOfProofInCivilJudgments}
Hao Li. 2008.
\newblock The allocation of burden of proof in civil judgments: An analysis based on the *gazette* case sample.
\newblock \emph{Tsinghua Law Review}, 6.

\bibitem[{Liu et~al.(2024)Liu, Feng, Xue, Wang, Wu, Lu, Zhao, Deng, Zhang, Ruan et~al.}]{liu2024deepseek}
Aixin Liu, Bei Feng, Bing Xue, Bingxuan Wang, Bochao Wu, Chengda Lu, Chenggang Zhao, Chengqi Deng, Chenyu Zhang, Chong Ruan, et~al. 2024.
\newblock Deepseek-v3 technical report.
\newblock \emph{arXiv preprint arXiv:2412.19437}.

\bibitem[{Merryman and P{\'e}rez-Perdomo(2018)}]{merryman2018civil}
John Merryman and Rogelio P{\'e}rez-Perdomo. 2018.
\newblock \emph{The civil law tradition: an introduction to the legal systems of Europe and Latin America}.
\newblock Stanford University Press.

\bibitem[{Ming(2021)}]{1021147323.nh}
Ni~Ming. 2021.
\newblock A study on the second-instance ruling in china's civil procedure.
\newblock Master's thesis, Soochow University.

\bibitem[{Qin et~al.(2024)Qin, Cao, Yu, Si, Chen, and Xu}]{qin2024explicitly}
Weicong Qin, Zelin Cao, Weijie Yu, Zihua Si, Sirui Chen, and Jun Xu. 2024.
\newblock Explicitly integrating judgment prediction with legal document retrieval: A law-guided generative approach.
\newblock In \emph{Proceedings of the 47th International ACM SIGIR Conference on Research and Development in Information Retrieval}, pages 2210--2220.

\bibitem[{Santosh et~al.(2022)Santosh, Xu, Ichim, and Grabmair}]{santosh2022deconfounding}
Tyss Santosh, Shanshan Xu, Oana Ichim, and Matthias Grabmair. 2022.
\newblock Deconfounding legal judgment prediction for european court of human rights cases towards better alignment with experts.
\newblock In \emph{Proceedings of the 2022 Conference on Empirical Methods in Natural Language Processing}, pages 1120--1138.

\bibitem[{Seidman(2016)}]{seidman2016new}
Guy~I Seidman. 2016.
\newblock The new comparative civil procedure.
\newblock In \emph{The Dynamism of Civil Procedure-Global Trends and Developments}, pages 19--44. Springer.

\bibitem[{Sergot et~al.(1986)Sergot, Sadri, Kowalski, Kriwaczek, Hammond, and Cory}]{sergot1986british}
Marek~J. Sergot, Fariba Sadri, Robert~A. Kowalski, Frank Kriwaczek, Peter Hammond, and H~Terese Cory. 1986.
\newblock The british nationality act as a logic program.
\newblock \emph{Communications of the ACM}, 29(5):370--386.

\bibitem[{Sun et~al.(2024)Sun, Huang, and Wei}]{sun2024chinese}
Jingyun Sun, Shaobin Huang, and Chi Wei. 2024.
\newblock Chinese legal judgment prediction via knowledgeable prompt learning.
\newblock \emph{Expert Systems with Applications}, 238:122177.

\bibitem[{Tong et~al.(2024)Tong, Yuan, Zhang, and Li}]{tong2024legal}
Suxin Tong, Jingling Yuan, Peiliang Zhang, and Lin Li. 2024.
\newblock Legal judgment prediction via graph boosting with constraints.
\newblock \emph{Information Processing \& Management}, 61(3):103663.

\bibitem[{Wiratunga et~al.(2024)Wiratunga, Abeyratne, Jayawardena, Martin, Massie, Nkisi-Orji, Weerasinghe, Liret, and Fleisch}]{wiratunga2024cbr}
Nirmalie Wiratunga, Ramitha Abeyratne, Lasal Jayawardena, Kyle Martin, Stewart Massie, Ikechukwu Nkisi-Orji, Ruvan Weerasinghe, Anne Liret, and Bruno Fleisch. 2024.
\newblock Cbr-rag: case-based reasoning for retrieval augmented generation in llms for legal question answering.
\newblock In \emph{International Conference on Case-Based Reasoning}, pages 445--460. Springer.

\bibitem[{Wu et~al.(2020)Wu, Kuang, Zhang, Liu, Sun, Xiao, Zhuang, Si, and Wu}]{wu-etal-2020-de}
Yiquan Wu, Kun Kuang, Yating Zhang, Xiaozhong Liu, Changlong Sun, Jun Xiao, Yueting Zhuang, Luo Si, and Fei Wu. 2020.
\newblock \href {https://doi.org/10.18653/v1/2020.emnlp-main.56} {De-biased court`s view generation with causality}.
\newblock In \emph{Proceedings of the 2020 Conference on Empirical Methods in Natural Language Processing (EMNLP)}, pages 763--780, Online. Association for Computational Linguistics.

\bibitem[{Xi(2018)}]{Xi2018SecondInstanceReversalInCivilAndCommercialCases}
Yuemin Xi. 2018.
\newblock The standards and rules for second-instance ruling changes in civil and commercial cases.
\newblock \emph{Journal of China University of Political Science and Law}, (3):110--124.

\bibitem[{Xiao et~al.(2021)Xiao, Hu, Liu, Tu, and Sun}]{xiao2021lawformer}
Chaojun Xiao, Xueyu Hu, Zhiyuan Liu, Cunchao Tu, and Maosong Sun. 2021.
\newblock Lawformer: A pre-trained language model for chinese legal long documents.
\newblock \emph{AI Open}, 2:79--84.

\bibitem[{Yang et~al.(2023)Yang, Xiao, Wang, Zhang, Bian, Yin, Lv, Pan, Wang, Yan et~al.}]{yang2023baichuan}
Aiyuan Yang, Bin Xiao, Bingning Wang, Borong Zhang, Ce~Bian, Chao Yin, Chenxu Lv, Da~Pan, Dian Wang, Dong Yan, et~al. 2023.
\newblock Baichuan 2: Open large-scale language models.
\newblock \emph{arXiv preprint arXiv:2309.10305}.

\bibitem[{Yang et~al.(2025)Yang, Li, Yang, Zhang, Hui, Zheng, Yu, Gao, Huang, Lv et~al.}]{yang2025qwen3}
An~Yang, Anfeng Li, Baosong Yang, Beichen Zhang, Binyuan Hui, Bo~Zheng, Bowen Yu, Chang Gao, Chengen Huang, Chenxu Lv, et~al. 2025.
\newblock Qwen3 technical report.
\newblock \emph{arXiv preprint arXiv:2505.09388}.

\bibitem[{Yang et~al.(2024)Yang, Yang, Zhang, Hui, Zheng, Yu, Li, Liu, Huang et~al.}]{yang2024qwen25}
an~Yang, Baosong Yang, Beichen Zhang, Binyuan Hui, Bo~Zheng, Bowen Yu, Chengyuan Li, Dayiheng Liu, Fei Huang, et~al. 2024.
\newblock Qwen2.5 technical report.
\newblock \emph{arXiv preprint arXiv:2412.15115}.

\bibitem[{Yue et~al.(2023)Yue, Chen, Wang, Li, Shen, Liu, Zhou, Xiao, Yun, Huang et~al.}]{yue2023disc}
Shengbin Yue, Wei Chen, Siyuan Wang, Bingxuan Li, Chenchen Shen, Shujun Liu, Yuxuan Zhou, Yao Xiao, Song Yun, Xuanjing Huang, et~al. 2023.
\newblock Disc-lawllm: Fine-tuning large language models for intelligent legal services.
\newblock \emph{arXiv preprint arXiv:2309.11325}.

\bibitem[{Zhang(2012)}]{Zhang2012EfficiencyAndFairnessInCivilProcedure}
Weiping Zhang. 2012.
\newblock The value trade-off between efficiency and fairness in the revision of the *civil procedure law*.
\newblock \emph{Chinese Judiciary}, (6):29--34.

\bibitem[{Zhao(2025)}]{Zhao2025ExplanationSystemAndAuthorityLogic}
Zhichao Zhao. 2025.
\newblock The authoritative logic behind the explanation system: Defending active clarification.
\newblock \emph{Journal of China University of Political Science and Law}, (02):148--159.

\bibitem[{Zheng et~al.(2022)Zheng, Liu, and Sun}]{zheng2022lawrec}
Min Zheng, Bo~Liu, and Le~Sun. 2022.
\newblock Lawrec: automatic recommendation of legal provisions based on legal text analysis.
\newblock \emph{Computational Intelligence and Neuroscience}, 2022(1):6313161.

\bibitem[{Zheng et~al.(2024)Zheng, Zhang, Zhang, Ye, and Luo}]{zheng-etal-2024-llamafactory}
Yaowei Zheng, Richong Zhang, Junhao Zhang, Yanhan Ye, and Zheyan Luo. 2024.
\newblock \href {https://doi.org/10.18653/v1/2024.acl-demos.38} {{L}lama{F}actory: Unified efficient fine-tuning of 100+ language models}.
\newblock In \emph{Proceedings of the 62nd Annual Meeting of the Association for Computational Linguistics (Volume 3: System Demonstrations)}, pages 400--410, Bangkok, Thailand. Association for Computational Linguistics.

\bibitem[{ZhihaiLLM(2023)}]{wisdom2024modelcard}
ZhihaiLLM. 2023.
\newblock \href {https://huggingface.co/ZhihaiLLM/wisdomInterrogatory} {{W}isdom {I}nterrogatory model card}.

\bibitem[{Zhong et~al.(2020)Zhong, Xiao, Tu, Zhang, Liu, and Sun}]{zhong2020does}
Haoxi Zhong, Chaojun Xiao, Cunchao Tu, Tianyang Zhang, Zhiyuan Liu, and Maosong Sun. 2020.
\newblock How does nlp benefit legal system: A summary of legal artificial intelligence.
\newblock In \emph{Proceedings of the 58th Annual Meeting of the Association for Computational Linguistics}, pages 5218--5230.

\bibitem[{Zhou et~al.(2024)Zhou, Shi, Song, Yang, Jin, Guo, and Li}]{zhou2024lawgpt}
Zhi Zhou, Jiang-Xin Shi, Peng-Xiao Song, Xiao-Wen Yang, Yi-Xuan Jin, Lan-Zhe Guo, and Yu-Feng Li. 2024.
\newblock Lawgpt: A chinese legal knowledge-enhanced large language model.
\newblock \emph{arXiv preprint arXiv:2406.04614}.

\bibitem[{Zhu and Xiao(2020)}]{Zhu2020AppealRequestAndSecondInstanceRuling}
Yaqi Zhu and Feng Xiao. 2020.
\newblock The relationship between appeal requests and the method of second-instance rulings: Focusing on article 170, paragraph 1 of the *civil procedure law*.
\newblock \emph{Legal Application}, (11):126--133.

\end{thebibliography}

\clearpage
\newpage
\appendix
\section{Background on Second-Instance Trials and Appellate Reversal in China} \label{appendix:background}

\subsection{Structural Differences Between Trial Court Levels}

The second-instance trial is a core component of China’s judicial process. Article 171 of the Civil Procedure Law of the People’s Republic of China (2023 Amendment) sets out the statutory conditions and time limits for initiating the second-instance procedure, thereby safeguarding the right of appeal. From a legal theory perspective, the second instance refers to the process by which a higher court reviews the judgment or ruling of the first instance upon appeal \cite{Jiang1998ResearchAchievements}. It plays a crucial role in correcting errors, ensuring uniform application of the law, and protecting the legitimate rights of the parties, thus upholding judicial fairness \cite{Zhang2012EfficiencyAndFairnessInCivilProcedure}. In practice, the second-instance judgment is final and, once rendered, carries legal effect, maintaining the authority and stability of judicial decisions.

Compared with the first instance, the second instance differs in several key aspects. In terms of procedure, as shown in Figure \ref{fig:fig1}, the first instance involves the initial trial, following steps such as filing, acceptance, trial, and judgment. In contrast, the second instance is initiated by appeal and focuses on reviewing errors in factual findings, legal application, and procedural legality of the first-instance judgment \cite{Li2008BurdenOfProofInCivilJudgments}. The second-instance court undertakes a comprehensive review of the entire first-instance process to ensure fairness. For example, in complex civil contract disputes, the first-instance court may issue an incorrect judgment due to misunderstandings of contract clauses. The second-instance court, through thorough review and precise legal interpretation, can correct such errors and effectively protect contractual rights \cite{Zhu2020AppealRequestAndSecondInstanceRuling}.

Regarding the scope of review, the first-instance court examines the facts and legal relationships of the case comprehensively, aiming to uncover the truth and apply the law correctly. The second-instance court generally reviews within the scope of the party's appeal, but may also review the entire case if necessary. This arrangement respects the party's right to dispose of the case while ensuring the court's comprehensive control. For instance, if an appeal concerns only part of the judgment but the second-instance court discovers significant errors elsewhere, it may review the entire case to ensure fairness.

\subsection{Categories of Appellate Reversal Grounds}

According to Article 177 of the Civil Procedure Law of the People’s Republic of China (2023 Amendment), after a comprehensive review of the appealed case, the second-instance court may reverse, annul, or modify the judgment, or remand the case for retrial if it finds errors in factual findings, legal application, or unclear basic facts in the original judgment \cite{Zhao2025ExplanationSystemAndAuthorityLogic}. Thus, the main grounds for reversal in the second instance focus on two areas: factual determination and legal application.

Factual determination errors refer to mistakes made by the original court in identifying or understanding key facts, such as erroneous classifications, unclear findings, or insufficient evidentiary support. Specifically, this includes circumstances where the evidence underlying the judgment is unreliable, insufficient, contradictory, or legally excluded, resulting in incorrect determination of the case’s nature or the allocation of rights and obligations \cite{Bodenheimer1987Jurisprudence}. Legal application errors include misjudgment of the legal nature (e.g., misclassifying a tort dispute as a contract dispute), improper citation of provisions (e.g., applying a general law where a special law should apply), or violations of the principle of non-retroactivity, all of which undermine the legal basis of the judgment.

Scholars have examined these two aspects in depth. In terms of factual determination, research shows that new evidence often reshapes the factual understanding of a case, significantly impacting judgment outcomes \cite{Xi2018SecondInstanceReversalInCivilAndCommercialCases}. When new evidence is sufficient to overturn the findings of the first-instance judgment, the probability of reversal increases substantially. Unclear factual determinations are also a major factor in reversals, as ambiguity or errors in key facts undermine the factual basis of the first-instance judgment. For legal application, misinterpretation or improper application of legal provisions directly affects the legality and fairness of the judgment, commonly leading to reversal. As Ni Ming noted, factual determination is the logical foundation for legal application, and the reasoning paths for factual and legal issues differ fundamentally in civil procedure \cite{1021147323.nh}. For example, whether a contract is established is a matter of factual determination, while the contractual nature falls under legal application. Collectively, these studies highlight the decisive role of factual determination and legal application in appellate reversals from multiple perspectives.

\section{Implementation of Tasks} \label{appendix:task}

In this section, we provide the instruction and output examples for the five tasks.

\subsection{Judgment Reversal Prediction from the first-instance perspective}

\begin{framed}
\footnotesize

\textbf{Instruction:} 

Read the following first-instance judgment and determine whether the appellate court will change the judgment if an appeal is filed. If it will, specify whether the reason for changing the judgment is an error in fact-finding or an error in the application of the law.

\#\# First-instance Documents

Civil Judgment of the People’s Court of Panyu District, Guangzhou City, Guangdong Province (2020) Yue 0113 Minchu No. 19865 (...omitted here...) If you do not accept this judgment, you may submit a notice of appeal to this court within fifteen days from the date of service of the judgment, providing copies according to the number of opposing parties, and appeal to the Guangzhou Intermediate People’s Court.

\#\# Second-instance Claims

Ma Zhuojun’s appeal requests: 1. To revoke the first-instance judgment and change the judgment to dismiss all of Zhai Yongdong’s claims; 2. To have Zhai Yongdong bear all the case acceptance fees.

\#\# Reason for reversal

Directly output the answer by selecting from ["Non-Reversed", "Factual determination errors", "Legal application errors"]. There may be multiple answers.

Do not output anything else.

\textbf{Output:} 

Non-Reversed

\end{framed}

\subsection{Judgment Reversal Prediction from the second-instance perspective}

\begin{framed}
\footnotesize

\textbf{Instruction:} 

Read the following first-instance judgment and determine whether the appellate court will reverse the judgment if an appeal is filed. If it will, specify whether the reason for changing the judgment is an error in fact-finding or an error in the application of the law.

\#\# First-instance Documents

Civil Judgment of Qianjiang District People’s Court, Chongqing City (2020) Yu 0114 Minchu No. 8176 (...omitted here...) If you do not accept this judgment, you may submit a notice of appeal to this court within fifteen days from the date the judgment is served, providing copies according to the number of opposing parties, and appeal to the Fourth Intermediate People’s Court of Chongqing City.

\#\# Second-instance Claims

Hou Zhangqing’s appeal requests: 1. To revoke the first-instance judgment and change the judgment to support the appellant’s claims in the first instance; 2. To have the appellee bear the case acceptance fees for both the first and second instance.

\#\# Second-instance Fact Description

\end{framed}

\begin{framed}
\footnotesize

The first-instance court’s finding that the appellant had no evidence to prove that he was injured while working on Building 3, Section 1 of the Banlishan Resettlement Housing Project is contrary to objective facts and legal provisions. (...omitted here...) This court confirms the other facts found in the first instance.

\#\# Reason for reversal

Directly output the answer by selecting from ["Non-Reversed", "Factual determination errors", "Legal application errors"]. There may be multiple answers.

Do not output anything else.

\textbf{Output:} 

Factual determination errors, Legal application errors

\end{framed}

\subsection{Legal Provision Recommendation}

\begin{framed}
\footnotesize

\textbf{Instruction:} 

You are a second-instance judge. Select the relevant legal provisions from the candidate statutes related to this case. There may be multiple applicable statutes. Please directly output the letter codes of the statutes in the answer section.

\#\# Second-instance Claims \& Fact Description

Civil Judgment of Jinzhou Intermediate People’s Court, Liaoning Province (2023) Liao 07 Min Zhong No. 795 Appellant (original defendant)... This court, after trial, found that the facts are basically consistent with the facts established by the first-instance judgment, which this court confirms.

\#\# Candidate Provisions

A. Article 61 of the Construction Law of the People’s Republic of China

B. Article 11 of the Supreme People’s Court Interpretation on Several Issues Concerning the Application of Law in Personal Injury Cases

(...omitted here...)

\#\# Answer

Directly output the letter codes only, do not output anything else.

\textbf{Output:} 

FGH

\end{framed}

\subsection{Legal Judgment Prediction}

\begin{framed}
\footnotesize

\textbf{Instruction:} 

You are a judge in a second-instance trial. Based on the facts established in the first-instance judgment, as well as the claims and facts presented in the second-instance appeal, determine whether the given claim should be upheld.

\#\# First-instance Facts

On August 1, 2011, both parties signed a real estate lease contract, under which the plaintiff agreed to lease part of the premises located on the east side of Building X, Haidian District, Beijing, to Party B for use (...omitted here...) During the litigation, Qianghua Printing Factory filed an application for property preservation with this court, requesting the preservation of Yinghao Hotel's property, and this court issued Civil Ruling (2021) Jing 0108 Minchu No. 69144.

\#\# Second-instance Claims \& Facts

Yinghao Hotel’s appeal requests: 1. To lawfully revoke items 2, 3, 5, and 7 of the first-instance judgment; 

\end{framed}

\begin{framed}
\footnotesize

\noindent (...omitted here...) This court finds as follows: The facts found in the first instance are correct. In addition, it is found that both parties have confirmed the payment obligations for the water and electricity fees involved in item 4 of the first-instance judgment, and Yinghao Hotel has already fulfilled them.

\#\# Claim

The plaintiff requests the court to order the defendant to vacate the two-story annex structure attached to the office building on the east side of Building X, Haidian District, Beijing (i.e., the leased property under the "Military Real Estate Lease Contract" signed by both parties, with a contract area of 534 square meters and an actual measured area of 811.04 square meters).

\#\# Claim support prediction

Directly output the answer by selecting one from ["Fully Support", "Partially Support", "Not Support"]. 

Do not output anything else.

\textbf{Output:} 

Fully Support
\end{framed}

\subsection{Court View Generation}

\begin{framed}
\footnotesize

\textbf{Instruction:} 

You are a judge in a second-instance trial. Based on the following facts, complete the court’s opinion section in the second-instance judgment (including the paragraphs starting with "The Court holds" and "The judgment is as follows").

\#\# Claims \& Fact Description

Civil Judgment of Jinzhou Intermediate People’s Court, Liaoning Province (2023) Liao 07 Min Zhong No. 795 Appellant (original defendant)... This court, after trial, found that the facts are basically consistent with the facts established by the first-instance judgment, which this court confirms.

\#\# Second-instance Claims \& Facts

Appellant Lvyuancheng Company’s appeal requests: 1. To request the second-instance court to lawfully revoke the original judgment, and lawfully change the judgment to dismiss the appellee’s claims in the first instance (...omitted here...) to deliver the relevant documents for handling the transfer registration of the commercial housing to the plaintiff; if the defendant delays delivery, it should bear liability for breach of contract.

\#\# Court’s Opinion

In this section, please directly output the court’s opinion (including the paragraphs beginning with "The Court holds" and "The judgment is as follows"). Do not output any other content.

The court holds...

\textbf{Output:} 

The court holds that although both parties agreed on the calculation standard for liquidated damages in the contract, the plaintiff did not provide evidence to prove the actual losses caused by the delay in handling the certificate (...omitted here...) 

In summary, in accordance with Article 170, Paragraph 1, Item 2 of the Civil Procedure Law of the People’s Republic of China, the judgment is as follows: 1. Revoke the Civil Judgment No. 1164 (2020) Qian 2723 Minchu of Guiding County People’s Court, Guizhou Province (...omitted here...) This judgment is a final judgment.

\end{framed}

\begin{table}[h]
    \footnotesize
    \centering
    \scalebox{0.9}{
    \begin{tabular}{cc}
    \toprule
    \textbf{Provider} & \textbf{Model}\\
    \midrule
    \multirow{4}{*}{DeepSeek} & DeepSeek-V3-0324 \cite{liu2024deepseek} \\
    & DeepSeek-R1 \cite{guo2025deepseek} \\
    & R1-Distill-Qwen-32B \cite{guo2025deepseek}\\
    & R1-Distill-Qwen-7B \cite{guo2025deepseek} \\
    \midrule
    \multirow{2}{*}{OpenAI} & gpt-4.1-2025-04-14 \tablefootnote{\href{https://openai.com/index/gpt-4-1}{https://openai.com/index/gpt-4-1}} \\
    & o4-mini-2025-04-16 \tablefootnote{\href{https://openai.com/index/introducing-o3-and-o4-mini}{https://openai.com/index/introducing-o3-and-o4-mini}} \\
    \midrule
    \multirow{5}{*}{Alibaba} & Qwen2.5-72B-Instruct \cite{yang2024qwen25} \\
    & QwQ-32B \tablefootnote{\href{https://huggingface.co/Qwen/QwQ-32B}{Qwen/QwQ-32B}} \\
    & Qwen3-32B-Instruct \cite{yang2025qwen3} \\
    & Qwen3-8B-Instruct \cite{yang2025qwen3}\\
    & Qwen2.5-7B-Instruct \cite{yang2024qwen25} \\
    \midrule
    \multirow{2}{*}{Meta} & LLaMA-3.3-70B-Instruct \tablefootnote{\href{https://huggingface.co/meta-llama/Llama-3.3-70B-Instruct}{meta-llama/Llama-3.3-70B-Instruct}} \\
    & LLaMA-3.1-8B-Instruct \tablefootnote{\href{https://huggingface.co/meta-llama/Llama-3.1-8B-Instruct}{meta-llama/Llama-3.1-8B-Instruct}} \\
    \midrule
    \multirow{2}{*}{ZhipuAI} & GLM-4-Air \cite{glm2024chatglm} \\
    & GLM-Z1-Air \cite{glm2024chatglm} \\
    \midrule
    xAI & Grok-3-mini \tablefootnote{\href{https://x.ai/news/grok-3}{https://x.ai/news/grok-3}} \\
    \midrule
    Baichuan & Baichuan2-7B-Chat \cite{yang2023baichuan} \\
    \midrule
    ByteDance & Doubao-1.5-pro \tablefootnote{\href{https://seed.bytedance.com/en/special/doubao_1_5_pro}{https://seed.bytedance.com/en/special/doubao\_1\_5\_pro}} \\
    \midrule
    Fudan Univ. & DISC-LawLLM \cite{yue2023disc} \\
    \midrule
    Zhejiang Univ. & Wisdom Interrogatory \cite{wisdom2024modelcard} \\
    \bottomrule
    \end{tabular}
    }
    \caption{Models and model providers involved in this paper.}
    \label{tab:provider}
\end{table}

\section{Experiment Details} \label{appendix:experiment}

\subsection{Settings of LLMs} \label{appendix:models}

We conducted experiments on 20 models from 10 providers, as shown in Table \ref{tab:provider}. All LLMs used the same default system prompt, "You are a helpful assistant," and the temperature and top\_p were set to each model’s default parameters.

\subsection{Settings of Fine-tuning}

We used 80\% of the AppealCase data to train the bert-base-chinese \cite{devlin-etal-2019-bert} and qwen3-0.6b \cite{yang2025qwen3} models for judgment reversal prediction.

We trained a LabelClassification layer for bert-base-chinese for multi-label classification, using the AdamW optimizer with a learning rate of $2 \times 10^{-5}$, a total batch size of 64, and 10 training epochs.

We performed full fine-tuning of Qwen3-0.8B using the llama-factory framework \cite{zheng-etal-2024-llamafactory}, with DeepSpeed ZeRO-3 and the AdamW optimizer, and a learning rate of $1 \times 10^{-5}$. The total batch size was set to 16 (4 batches per GPU), and the training lasted for 2 epochs.

\section{Case Study on Error Analysis}

In the Judgment Reversal Prediction task, among the \numprint{10000} cases in AppealCase, there are 401 cases for which all 20 models produced incorrect predictions. All of these 401 cases are cases of judgment reversal, and among them, 83\% were reversed due to errors in the application of law.

We conducted an analysis of these cases and identified the following main challenges:

\begin{itemize}[itemsep=3pt,topsep=3pt,parsep=0pt]

\item Dynamic legal knowledge: For example, regarding the determination of whether the monthly interest rate exceeds four times the one-year loan market quotation rate published by the National Interbank Funding Center, the benchmark interest rate here is dynamic. The models may not be able to access or accurately understand the relevant data in time, leading to prediction errors.

\item Easily confused legal relationships: In some cases, there are legal relationships that are easily confused, such as mistakenly identifying a lending relationship as a partnership, or an employment relationship as a contract-for-work relationship. Due to the complexity of the facts or unclear statements, the models are prone to bias in legal relationship judgment, which affects the final prediction results.

\item Issues in the division of liability: In some cases, the division of liability among parties is more complex, and the appellate court has adjusted the proportion or method of liability compared with the first-instance judgment. The models struggle to accurately capture the details of liability determination and the judicial reasoning, resulting in prediction deviations.
\end{itemize}

These hard cases reflect the current limitations of models in handling complex cases and represent a key direction for improving predictive accuracy.

\end{document}